\documentclass[preprint,review,12pt]{elsarticle}

\usepackage{amssymb}
\usepackage{multirow} 
\usepackage{booktabs}
\usepackage{tabularx}
\usepackage{array}
\usepackage{subcaption}
\usepackage{booktabs}
\usepackage{longtable}

\journal{Elsevier} 

\begin{document}

\begin{frontmatter}

\title{Depth Map Denoising Network and Lightweight Fusion Network for Enhanced 3D Face Recognition}

\author[1]{Ruizhuo Xu\fnref{fn1}}
\ead{ruizhuoxu@bupt.edu.cn}

\author[2]{Ke Wang\fnref{fn1}}
\ead{wangkeai@chinamobile.com}

\author[2]{Chao Deng}
\ead{dengchao@chinamobile.com}

\author[1]{Mei Wang}
\ead{wangmei1@bupt.edu.cn}

\author[2]{Xi Chen}
\ead{chenxiyjy@chinamobile.com}

\author[2]{Wenhui Huang}
\ead{huangwenhui@chinamobile.com}

\author[2]{Junlan Feng}
\ead{fengjunlan@chinamobile.com}

\author[1]{Weihong Deng\corref{cor1}}
\ead{whdeng@bupt.edu.cn}

\cortext[cor1]{Corresponding author}
\fntext[fn1]{These authors contributed equally to this work and should be
  considered co-first authors.}

\affiliation[1]{organization={School of Artificial Intelligence},
            addressline={Beijing University of Posts and Telecommunications}, 
            city={Beijing},
            postcode={100876}, 
            country={China}}
\affiliation[2]{organization={China Mobile Research Institute},
            city={Beijing},
            postcode={100053}, 
            country={China}}

\begin{abstract}
With the increasing availability of consumer depth sensors, 3D face recognition (FR)
has attracted more and more attention. However, the data acquired by these sensors
are often coarse and noisy, making them impractical to use directly.
In this paper, we introduce an innovative \textit{Depth map denoising network (DMDNet)} based on
the \textit{Denoising Implicit Image Function (DIIF)} to reduce noise and enhance
the quality of facial depth images for low-quality 3D FR.
After generating clean depth faces using DMDNet,
we further design a powerful recognition network
called \textit{Lightweight Depth and Normal Fusion network (LDNFNet)},
which incorporates a \textit{multi-branch fusion block} to learn unique and complementary
features between different modalities such as depth and normal images.
Comprehensive experiments conducted on four distinct low-quality databases demonstrate the
effectiveness and robustness of our proposed methods.
Furthermore, when combining DMDNet and LDNFNet, we achieve state-of-the-art results on
the Lock3DFace database.
\end{abstract}

\begin{graphicalabstract}
\end{graphicalabstract}

\begin{highlights}
\item A novel 3D face denoising network based on the implicit neural representation
\item Positonal encoding and multi-scale decoding fusion strategy help to denoise
\item A lightweight fusion network achieving high face recognition performance
\end{highlights}


\end{frontmatter}


\section{Introduction}
\label{}
Owing to technological breakthroughs in deep learning
\cite{szegedy2015going,he2016deep}
and the significant advancement in 2D face recognition (FR) \cite{li2018distance,wei2020minimum,huang2022deep}, 
3D FR has drawn increasing attention in the computer vision community.
Especially in 2017, the release of iPhoneX marked the first year of the application of 3D FR
in smartphone scenarios, which showcased the enormous potential among academic and industry communities.

Unlike 2D face images, 3D face data provides additional information
such as the geometric structure of the face, including depth, shape, and curvature.
This extra information not only allows for more accurate face recognition
but also facilitates reliable and robust face anti-spoofing by detecting impersonation attacks.
Moreover, 3D faces have a certain degree of privacy preservation,
which can effectively prevent biological information leakage from being used for the black industry chain.
In the last two decades, some 3D face databases have been proposed for academic research,
such as FRGC v2 \cite{phillips2005overview},
Bosphorus \cite{savran2008bosphorus}, and BU3DFE \cite{yin20063d}.
However, most of them are collected by expensive and sophisticated 3D scanners,
making them unaffordable for practical applications.
Recently, consumer depth cameras such as Kinect and RealSense have become increasingly
popular, allowing users to acquire 3D data at an affordable price.
This has significantly propelled the development of the 3D vision field \cite{tu2023consistent,zhang2022uncertainty,chen2021joint}.
Nevertheless, depth images or point clouds collected by these devices are relatively rough and noisy,
presenting a huge challenge for downstream tasks.
As a result, developing ways to obtain clean and smooth faces using low-quality depth face
data is crucial for 3D FR.



\begin{figure}[!t]
 \centering 
 \includegraphics[width=5.0in]{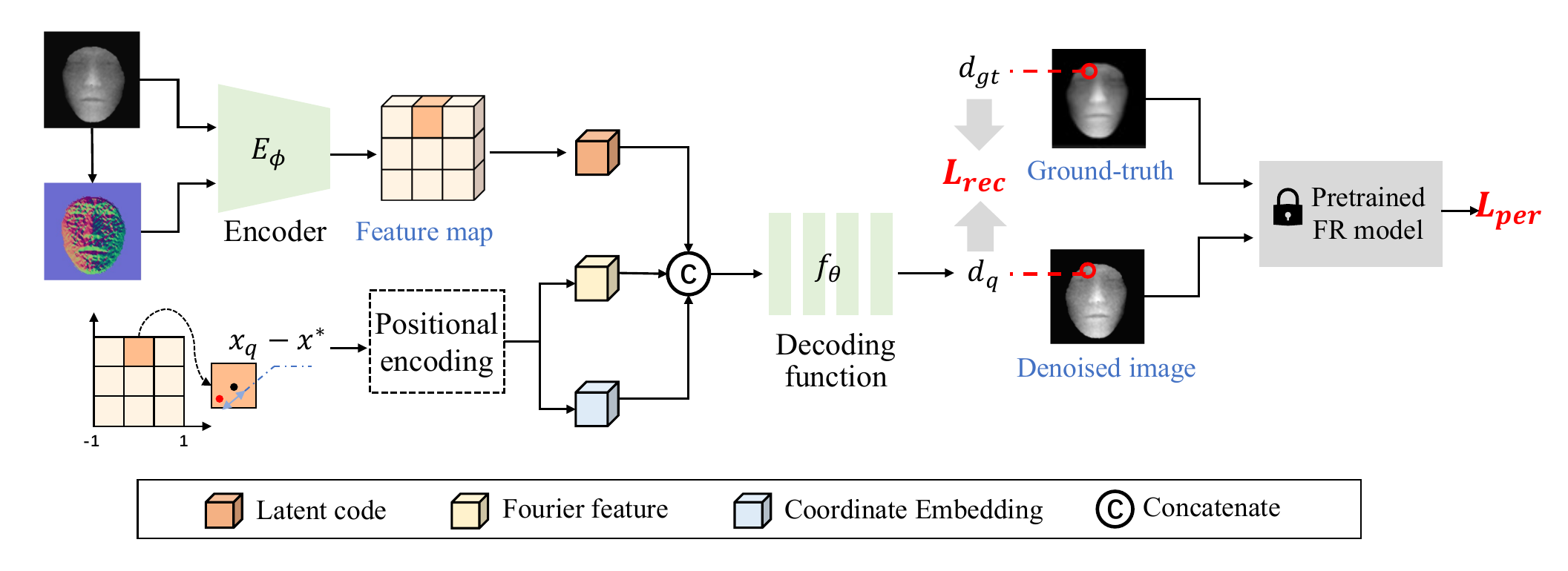}
 \caption{
  The overall structure of the proposed DMDNet. Our DMDNet consists of an
  encoder $E_{\phi}$ and a decoding function $f_{\theta}$. $E_{\phi}$
  represents noisy depth image as a set of latent codes distributed in
  2D spatial domain, while the $f_{\theta}$ takes the coordinate information
  and the latent code queried by the coordinate as input and predicts the
  denoised depth value at the given coordinate.
 }
 \label{fig1}
\end{figure}

In this paper, we view face denoising from the perspective of
\textit{implicit neural representation} and propose a novel
\textit{Depth Map Denoising Network (DMDNet)} based on the \textit{Denoising Implicit Image Function (DIIF)}  
to remove noises and improve the quality of facial depth images for low-quality 3D face recognition.
Our DMDNet comprises an encoder and a decoding function, as illustrated in Figure \ref{fig1}.
The encoder represents noisy depth images as a set of latent codes
distributed in spatial dimensions, while the decoding function takes
the coordinate information and the latent code queried by the coordinate
as input and predicts the denoised depth value at the given coordinate.
Since face images are highly structured, our DMDNet utilizes the spatial and
semantic information implicit in the coordinate to provide a strong prior for denoising.
Compared to traditional denoising networks based on an encoder-decoder approach,
our DMDNet achieves better results.
Moreover, we also leverage \textit{positional encoding} to encode the coordinate information
such that the denoising performance in high-frequency details can be improved,
and simultaneously, we propose a \textit{Multi-Scale Decoding Fusion} strategy
which jointly utilize latent codes from different levels for decoding,
achieving robustness to the change of spatial structure and improving the representation ability.

To achieve better 3D FR performance using these denoised images,
we further design a novel and efficient recognition network,
called \textit{Lightweight Depth and Normal Fusion network (LDNFNet)},
which takes advantage of multiple modalities (i.e. depth and normal images) for recognition.
In addition to learning specific features from every single modality,
we design a multi-branch fusion block to learn common features between
different modalities, in which the features from different modalities are fused
in an efficient way.

To summarize, the contributions of this paper are as follows:
\begin{itemize}
\item{To the best of our knowledge, this is the first work that
implicit neural representation is introduced into 3D face denoising.
The utilization of coordinate information provides a strong prior for denoising.
Additionally, positional encoding and multi-scale decoding fusion strategies
help to obtain better denoising performance.}
\item{We propose a novel LDNFNet for 3D face recognition,
in which the multi-branch fusion block achieves the modal fusion
between depth and normal images.
It simultaneously learns specific and common features while reducing the extra computational overhead.}
\item{Our proposed DMDNet outperforms existing methods on 3D face denoising tasks.
When combining DMDNet and LDNFNet, we achieve state-of-the-art results
on Lock3DFace datasbase.}
\end{itemize}

\section{Related Works}
In this section, we concisely review the related methods of depth map denoising, low-quality 3D FR,
and Implicit Neural Representations (INRs).

\subsection{Depth Map Denoising}
Traditional depth map denoising methods aim to generate high-quality depth maps
by fusing multiple consecutive low-quality depth frames.
For instance, KinectFusion \cite{newcombe2011kinectfusion} presents
a real-time reconstruction pipeline for indoor rigid scenes
using consumer depth sensors. DynamicFusion \cite{newcombe2015dynamicfusion} extends this goal to non-rigid scenes.
While fusion-based methods effectively reduce noise
and enhance the quality of the original depth maps,
they often involve complexity and require sequences of depth maps as input,
resulting in a certain time delay.

In recent years, several studies have explored the use of neural networks for denoising
and refining low-quality depth maps. DDRNet \cite{yan2018ddrnet} proposes a cascaded depth denoising and refinement
network that enhances the quality of noisy depth maps by leveraging multi-frame fused geometry
and high-quality color images through joint training.
DDRNet shows promising results for various generic 3D objects.
However, it neglects to address the need for identity consistency in the specific task of 3D face denoising.

On the other hand, 3D-FRM \cite{mu2021refining} presents an innovative and lightweight 3D face refinement model
to reduce noise in low-quality facial depth maps.
It improves face recognition performance on the Lock3DFace database using the denoised depth face images.
The structure of 3D-FRM is based on a fully convolutional encoder-decoder architecture.
Although convolutional neural networks possess many excellent properties like parameter sharing and translation invariance,
they overlook the coordinate information in the depth image,
which is helpful for the network to perceive the spatial structure of the face.

To address these limitations, we propose a novel Depth Map Denoising Network (DMDNet) based on implicit neural representation.
DMDNet explicitly leverages coordinate information and enhances denoising performance on depth face images.
Additionally, we incorporate a perceptual loss to ensure identity consistency before and after denoising.

\subsection{Low-quality 3D Face Recognition}
In the last few decades, 3D FR has received significant attention, especially with the
release of high-quality 3D face databases like FRGC v2 \cite{phillips2005overview},
BU3D-FE \cite{yin20063d}, and Bosphorus \cite{savran2008bosphorus}. 
High-quality 3D FR has demonstrated impressive recognition results,
but the performance of low-quality 3D FR remains unsatisfactory.

Some early works \cite{berretti2012superfaces,min2014kinectfacedb} primarily relied on
hand-crafted descriptors such as LGBP and HOG.
Despite promising results, these methods used limited subjects
and lacked generalizability to other databases.
To address this, Zhang et al. \cite{zhang2016lock3dface} first introduced a public comprehensive low-quality 3D face database
named Lock3DFace, comprising 5,671 video sequences of 509 individuals.
They presented a baseline recognition result using the traditional approach, i.e. ICP.
Subsequently, Cui et al. \cite{cui2018improving} established a baseline for the deep learning method
by applying Inception v2 \cite{ioffe2015batch} to Lock3DFace.
To further enhance recognition accuracy, Mu et al. \cite{mu2019led3d} designed a
lightweight yet efficient deep model called Led3D and proposed a data processing system
including point-cloud recovery, surface refinement, and data augmentation, resulting in finer and bigger training data.
Lin et al. \cite{lin2021high} employed pix2pix \cite{isola2017image} network to generate
high-quality faces from noisy ones. Additionally, they introduced a multi-quality fusion
network (MQFNet) to fuse data of different qualities and improve FR performance.
Zhao et al. \cite{zhao2022lmfnet} proposed a lightweight multiscale fusion network
(LMFNet) with a hierarchical structure, achieving superior performance on Lock3DFace.

Among these methods, Led3D \cite{mu2019led3d} and MQFNet \cite{lin2021high} are most
relevant to our LDNFNet.
Led3D introduced an efficient deep model for 3D face recognition.
In our work, we extend Led3D into a three-path form to fully utilize the complementary information
between different modalities.
On the other hand, MQFNet uses general convolutional layers to achieve feature fusion
from data of different qualities.
However, we propose an innovative multi-branch convolutional fusion block to mine
more diverse fusion representations with fewer parameters and computations.

\subsection{Implicit Neural Representations}
Implicit neural representation is a widely used method for representing objects as
a multi-layer perceptron (MLP) that maps coordinates to signals in a specific domain.
This idea has been applied to modeling 3D objects \cite{park2019deepsdf,chen2019learning},
3D scene surface \cite{jiang2020local,mildenhall2021nerf}
and 2D images \cite{chen2021learning,sitzmann2020implicit}.

In the 3D domain, Chen et al. \cite{chen2019learning} proposed an implicit field decoder
that generates 3D shapes, which can be easily transferred to a variety of applications,
including generation, interpolation, and single-view reconstruction.
Deepsdf \cite{park2019deepsdf} is a representation of signed distance functions
(SDFs) of shapes via latent code conditioned feed-forward decoder networks, 
achieving state-of-the-art performance for learned 3D shape representation. 

In the 2D domain,
Sitzmann et al. \cite{sitzmann2020implicit} replaced the ReLU in the MLP with a
periodic activation function (sinusoidal) and demonstrated that it can model
more natural and refined images.
Chen et al. \cite{chen2021learning} proposed the Local Implicit Image Function (LIIF)
for representing images as a series of spatially distributed latent codes.
The decoding function takes the coordinate information and queries the latent code
around the coordinate as inputs, then predicts the RGB value at the given coordinate
as an output. Due to the continuous coordinates, LIIF can be presented in arbitrary resolution.

Referring to LIIF \cite{chen2019learning},
we propose Denoising Implicit Image Function (DIIF) to introduce the concept of
implicit neural representation into the field of depth map denoising.
In addition, we propose two effective strategies:
positional encoding and multi-scale decoding fusion (MSDF),
which greatly improve the performance of denoising and enhance the denoising metrics.

\section{Depth Map Denoising Network}

\begin{figure}[!t]
 \centering 
 \includegraphics[width=5.0in]{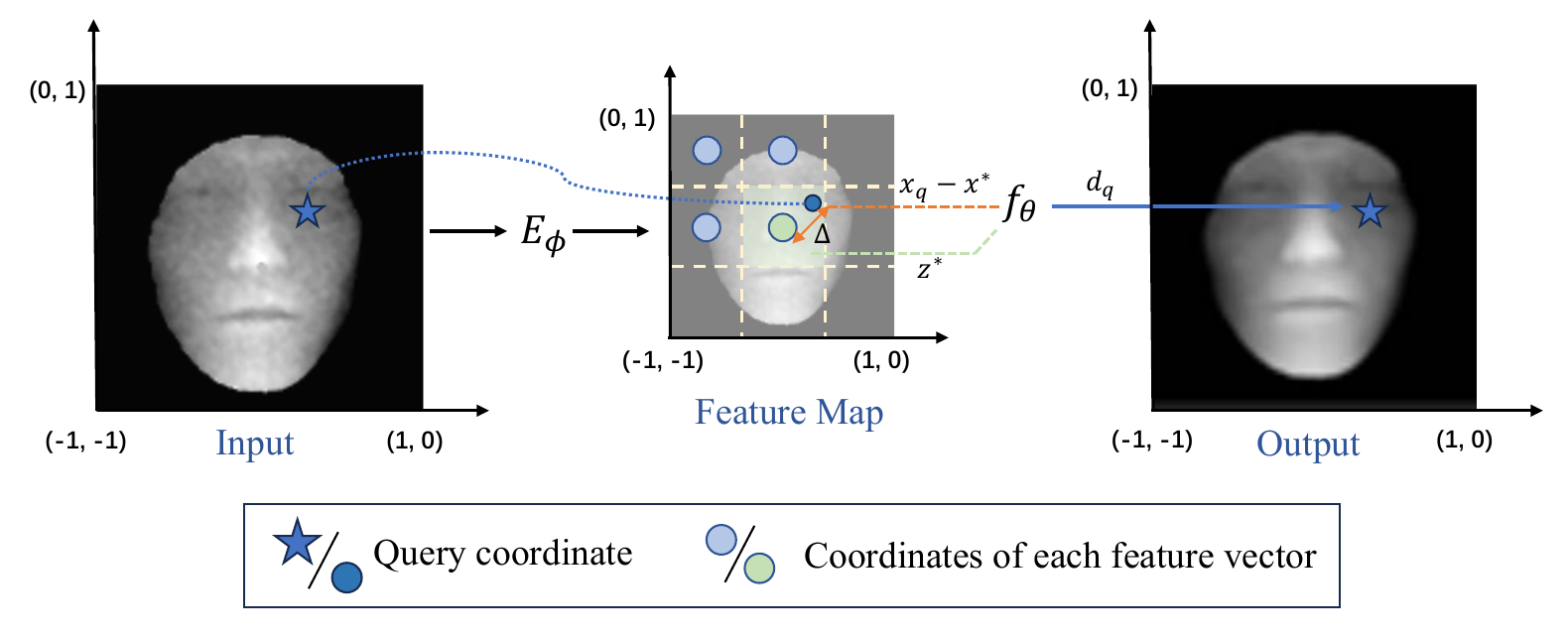}
 \caption{
  Illustration of DIIF. The input noisy depth map is represented as a feature map
  $F \in \mathbb{R}^{C \times H \times W}$
  by the encoder $E_{\phi}$. Each feature vector
  $z_{i} \in \mathbb{R}^{C \times 1 \times 1}$
  in the feature map
  (shown as a grid of yellow dashed lines in the middle plot) is assigned a coordinate.
  Given a query coordinate $x_{q}$, we first select the closest feature vector $z^{*}$
  (depicted as the green dot in the middle plot with coordinate $x^{*}$)
  based on the Euclidean distance from $x_{q}$ to the position of each feature vector,
  and use it as the latent code for this local region. Then, the decoding function $f_{\theta}$ takes latent code $z^{*}$
  and the distance vector ($x_{q} - x^{*}$) as inputs,
  and outputs predicted denoised depth values at $x_{q}$.
  }
 \label{fig:diif}
\end{figure}

In this paper, we propose a novel \textit{Depth Map Denoising Network (DMDNet)}
based on the \textit{Denoising Implicit Image Function (DIIF)}
to effectively remove noise on low-quality depth face images.
An overview of the network is demonstrated in Figure \ref{fig1},
which contains a fully convolutional encoder $E_{\phi}$
and a decoding function $f_{\theta}$.
$E_{\phi}$ used to extract the features $F \in \mathbb{R}^{C \times H \times W}$ of the input noisy depth map $D_{in}$.
$f_{\theta}$ is an MLP with parameters $\theta$ that takes coordinate
$x \in \mathbb{R}^{2}$ and corresponding feature vector
$z \in \mathbb{R}^{C \times 1 \times 1}$ located in the 2D spatial domain of $F$
as input and predicts the clean signal (i.e. the depth value).
In the following, we will provide a detailed explanation of \textit{DIIF}
and the specific structure of \textit{DMDNet}.

\subsection{Denoising Implicit Image Function}
Inspired by recent advances in implicit neural representations (INRs)
for 3D object/scene representation
\cite{park2019deepsdf,jiang2020local,sitzmann2020implicit} 
and 2D image generation \cite{sitzmann2020implicit,chen2021learning,anokhin2021image},
we propose Denoising Implicit Image Function representation
to explore the feasibility of leveraging INRs for depth map denoising.

Typically, INRs use an MLP to map coordinates to signals in a specific domain.
To share the knowledge across different input instances,
DIIF adopts the encoder-based approach
\cite{sitzmann2020implicit,chen2021learning},
where each instance is encoded as feature vectors (referred to as latent codes)
and concatenated with coordinates as input to the MLP.
Similar to works such as LIIF \cite{chen2021learning} and LIG \cite{jiang2020local},
DIIF is defined as:

\begin{equation}
  \label{eq1}
  d_{q} = f_{\theta}(z^{*},x_{q}-x^{*}),
\end{equation}
\begin{equation}
  \label{eq2}
  z^{*} = E_{\phi}(D_{in})[x^{*}],
\end{equation}
where $d_{q}$ is the clean depth value at coordinate $x_q$ on the denoised depth map, and
$z^{*}$ is a local latent code at coordinate $x^{*}$ in the feature map,
which is closest to $x_q$.
To ensure consistency, we normalize all coordinates to the range $[-1, 1]$,
enabling coordinates from different domains to share the same coordinate system.

An illustration of DIIF is presented in Figure \ref{fig:diif}.
Firstly, DIIF encodes the input noisy depth map into a feature map using an encoder network $E_{\phi}$.
This feature map can be seen as a grid of feature vectors, each associated with a spatial coordinate.
To denoise a query point $x_{q}$, DIIF selects the closest feature vector based on Euclidean distance,
which serves as the latent code representing the local region around the query point.
Subsequently, the decoding function $f_{\theta}$ takes this latent code $z^{*}$,
along with the distance vector ($x_{q} - x^{*}$) between the query point and the coordinate of the selected feature vector,
to predict the denoised depth value at the query location. 

\begin{figure}[!t]
 \centering 
 \includegraphics[width=5.0in]{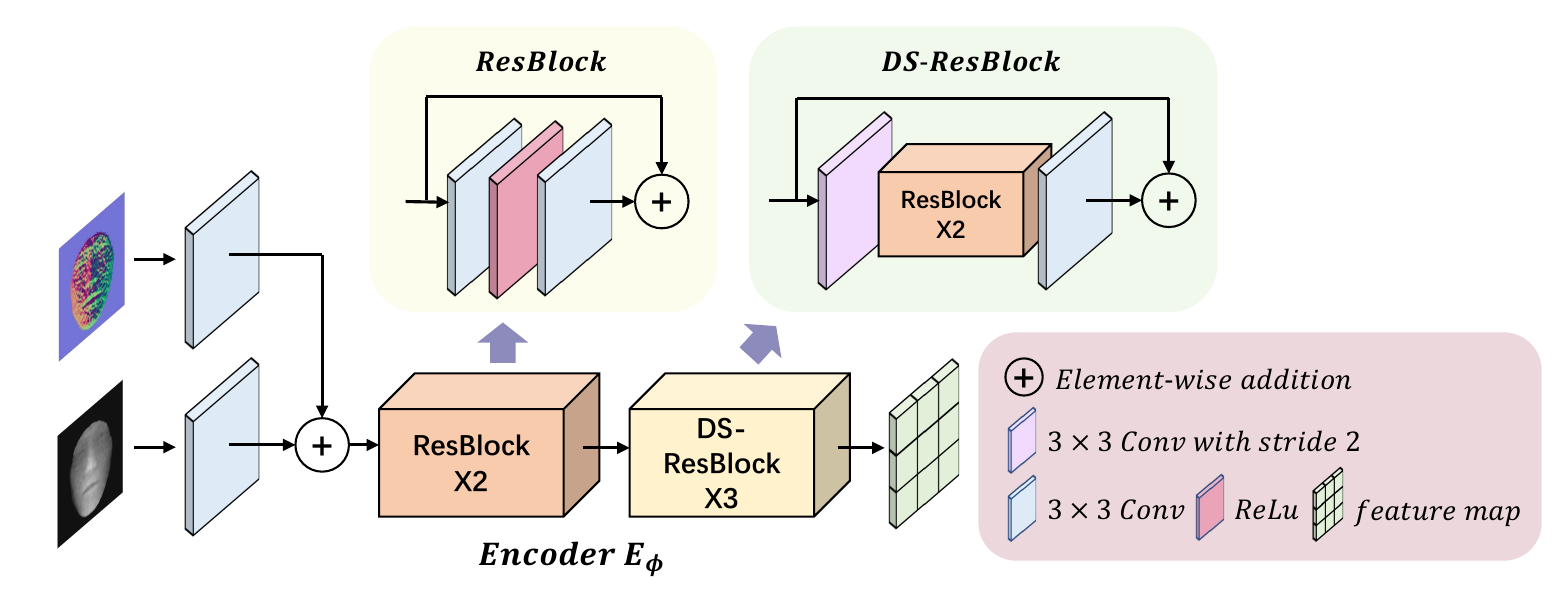}
 \caption{
  The detailed structure of the DMDNet encoder. The $3 \times 3$ convolutional
  layer here has a default stride of 1 and a padding of 1.
  The $\times N$ in the cuboid indicates the number of blocks.
  }
 \label{fig2}
\end{figure}
\subsection{Convolutional Encoder}\label{Encoder}
The architecture of the encoder network is shown in Figure \ref{fig2}.
Firstly, We compute the corresponding normal map from the depth map
and feed both into a convolutional layer with a kernel size of $3\times3$ respectively.
Then, we utilize a series of ResBlocks proposed in EDSR \cite{lim2017enhanced}
to extract high-level information from the combined features
while retaining the original information.
Furthermore, three DS-ResBlcoks are used to obtain feature maps at different resolutions.
Each DS-ResBlcok downsamples the feature map by a $3 \times 3$ convolutional layer
with a stride of 2 and then integrates the features using three ResBlcoks
with a $3 \times 3$ convolutional layer.
By extracting the output of the ResBlocks and each DS-ResBlcoks,
we can obtain four feature maps possessing varying resolutions (128, 64, 32, and 16) for use in the decoding stage.
A discussion on the design of our encoder is given in Appendix B.

\subsection{Decoding Function}
We employ a five-layer MLP with decreasing hidden dimensions
(256,\\128,64,32) as decoding function.
Each linear layer in the MLP is followed by ReLU activation,
except for the final layer, which uses Tanh activation to normalize output.
Although simply using an MLP as the decoding function can achieve impressive results,
its reconstruction performance in high-frequency details is still unsatisfactory.
Based on this, we apply two effective strategies to improve the decoding performance of the MLP,
namely \textit{positional encoding} and \textit{multi-scale decoding fusion}.

\textbf{Positional Encoding:}
In recent works \cite{sitzmann2020implicit,mildenhall2021nerf,tancik2020fourier},
it has been shown that incorporating coordinate position encoding can significantly enhance
reconstruction of high-frequency object details.
We adopt a combination of \textit{Fourier Features} and \textit{Coordinate Embeddings}, as employed in \cite{anokhin2021image,tancik2020fourier},
for our positional encoding.

Fourier Features is a linear layer with a $sine$ activation function
that maps the 2D coordinate vector $x$ to a N-dimensional Fourier vector
$e_{ff} \in \mathbb{R}^N$.
The Fourier Features encoding can be expressed as follows:
\begin{equation}
  \label{eq3}
  e_{ff} = sin(W_{ff}x_i),
\end{equation}

\begin{figure}[!t]
 \centering 
 \includegraphics[width=5.0in]{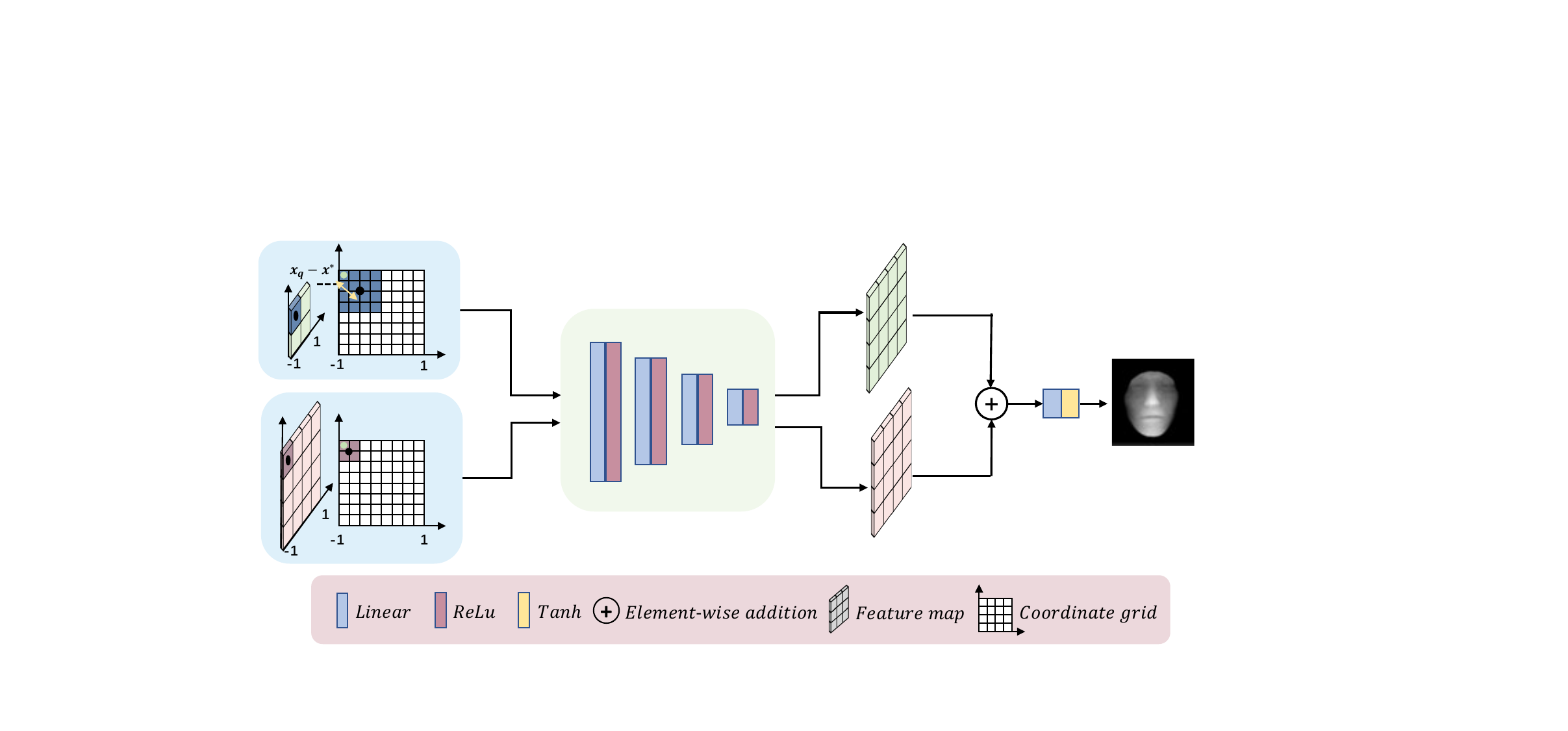}
 \caption{
  A simple illustration of MSDF. The feature maps obtained from the Encoder
  at different resolutions are fed into the decoding function in tune,
  along with the query coordinates. Their intermediate features are fused
  by element-wise addition before the last linear layer. Finally, the fused
  features are fed into the final linear layer with Tanh activation to predict
  the denoised depth values.
  }
 \label{fig3}
\end{figure}

where $x_i \in \mathbb{R}^2$ represents the initial coordinate vector,
$W_{ff} \in \mathbb{R}^{N\times2}$ is a learnable weight matrix shared by all coordinates,
and $sin(\cdot)$ denotes the sine activation function.

Coordinate Embeddings, on the other hand, train a distinct vector
$e_{ce}^i \in \mathbb{R}^N$ for each spatial coordinate.
This further improves the network's ability to reconstruct fine-grained details and can
even enable key points localization on faces.
The Coordinate Embeddings encoding can be written as follows:
\begin{equation}
  \label{eq4}
  e_{ce}^{i} = W_{ce}^{i}x_{i},
\end{equation}
where $W_{ce}^i \in \mathbb{R}^{N\times2}$ represents a trainable transformation matrix.
Each matrix is trained separately for each coordinate, thus allowing for more expressive 
representation of high-frequency information.


\textbf{Multi-Scale Decoding Fusion:}
Convolutional neural network is a stack of multiple convolutional layers
in which shallower layers usually have smaller receptive fields
to capture low-level features,
and deeper layers have larger receptive fields to capture abstract semantic features.
It is natural to combine features at different layers to reconstruct better images.
Motivated by this idea, we propose a Multi-Scale Decoding Fusion strategy (MSDF)
to make full use of the features at different scales of the encoder.
A simple illustration is shown in Figure \ref{fig3}.

In practice, we first selected four feature maps (mentioned in Section \ref{Encoder}) with different resolutions
from the encoder, corresponding to the information captured by different receptive fields.
We then feed these four feature maps into the decoding function $f_{\theta}$ in turn,
along with the query coordinates, and fuse their intermediate features
by element-wise addition before the last linear layer.
Finally, the fused features are fed into the final linear layer
with a Tanh activation function to obtain the denoised depth values.

Latent codes from feature maps at different resolutions
can represent decoding functions at multiple scales,
thereby greatly improving the MLP's expressive ability.
Moreover, we can notice that since the resolution of the query coordinate grid is the same,
and yet the resolutions of the feature maps are different,
the relative coordinates ($x_q-x^{*}$ in Equation \ref{eq1}) calculated
according to the nearest principle, are also different.
More precisely, the relative coordinate distance increases with the expansion of
the receptive field of the feature map, which helps the network to perceive the change of
spatial structure brought by different scale features.

\subsection{Loss Function}
The goal of DMDNet is to generate denoised depth faces while preserving identity information.
Therefore, we propose a joint loss function consisting of an L1 loss 
and SSIM \cite{wang2004image} loss for image reconstruction, 
and a perceptual loss for identity preservation.

We utilize a pre-trained face recognition model to compute the perceptual loss,
which can be formulated as:
\begin{equation}
  \label{eq6}
  L_{per}^{l_1} = \Vert F(D_{pred}) - F(D_{gt}) \Vert_1,
\end{equation}
where $F(\cdot)$ denotes the feature vector extracted by the recognition model,
$D_{pred}$ is the denoised depth map generated by DMDNet,
and $D_{gt}$ represents the supervised ground truth.

The overall loss function can be expressed as follows:
\begin{equation}
  \label{eq7}
  L = \lambda_{1}L_{rec}^{l_1} + \lambda_{2}L_{rec}^{SSIM} + \lambda_{3}L_{per}^{l_1},
\end{equation}
where parameters $\lambda_1$,$\lambda_2$, and $\lambda_3$ are used to balance the importance of different losses.

\begin{figure}[!t]
 \centering 
 \includegraphics[width=5.0in]{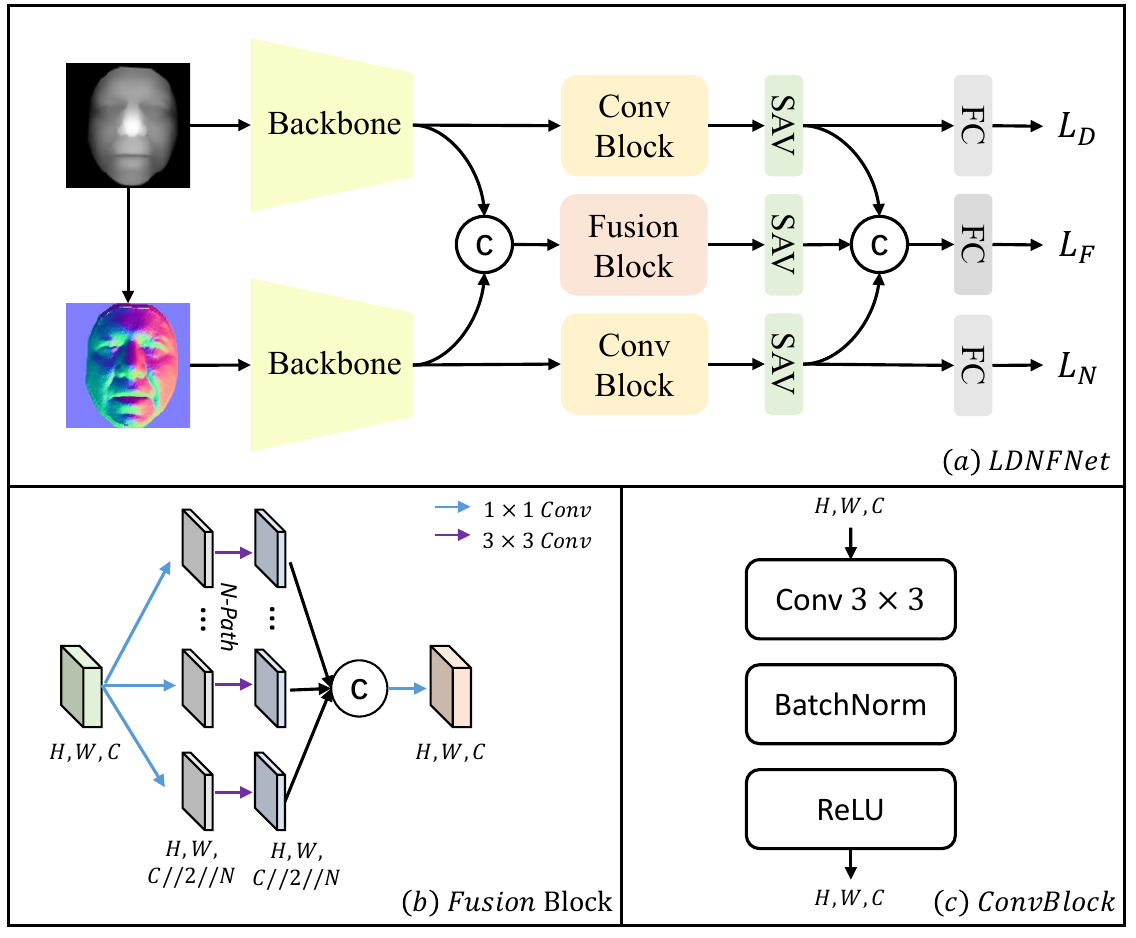}
 \caption{
  (a) Architecture of our LDNFNet. The structure of the Backbone
  is the same as Led3D \cite{mu2019led3d}.
  (b) Detailed structure of the Fusion Block. We take the multi-branch convolutional
  structure proposed in \cite{xie2017aggregated} to build the fusion block.
  (c) Detailed structure of the ConvBlock.
  }
 \label{fig:LDNFNet}
\end{figure}

\section{Lightweight Depth and Normal Fusion Network}

After recovering clean depth faces using DMDNet,
we propose a \textit{Light-\\weight Depth and Normal Fusion Network (LDNFNet)} for 3D face recognition.
The detailed structure of LDNFNet is shown in Figure \ref{fig:LDNFNet}, which contains three paths:
depth, normal, and fusion.
The depth and normal paths extract unique information for their respective modalities, while
the fusion path extracts complementary information between modalities using a lightweight
but efficient \textit{multi-branch fusion block}.

In the depth and normal paths, depth features and normal features are extracted separately
through a backbone network as Led3D \cite{mu2019led3d}, which contains 4 convolutional blocks (ConvBlocks).
Each block is composed of a convolutional layer with a kernel size of $3 \times 3$,
a batch normalization layer, and a ReLU activation layer.
The output features of each block are concatenated together in the channel dimension by the
\textit{MSFF} \cite{mu2019led3d} module. Additionally, feature maps at different scales
are integrated by another Convblock to generate more discriminative representations of 3D faces.
Finally, the integrated features of the depth map and normal map are vectorized by the SAV module \cite{mu2019led3d},
which is a spatial attention-weighted pooling of the feature maps.
We utilize the Fully-Connected (FC) layer and the Softmax layer with the cross-entropy loss
to compute the identification losses ($L_D$ and $L_N$) for the depth path and normal path.

In the fusion path, LDNFNet concatenates the depth features and normal features
outputted by the MSFF module in the channel dimension and obtains the fused features through a 
fusion block.
We used the multi-branch convolutional structure proposed in ResNext \cite{xie2017aggregated}
to build the fusion block shown in Figure \ref{fig3}b.
Each branch shares the same topology, consisting of a $1 \times 1$ convolutional layer
and a $3 \times 3$ convolutional layer, which is equivalent to projecting the input features into several different feature subspaces,
where the network can learn and discover more diverse feature representations.
In addition, this multi-branch operation can also be seen as a form of regularization.
With each additional branch, the network learns features with sparser relationships.
Although there is complementary information between depth maps and normal maps,
there is also a lot of redundant information.
Therefore, this sparsity greatly reduces the risk of overfitting.
Finally, the outputs of each branch are aggregated by concatenation and a $1 \times 1$ convolutional layer.
In practice, LDNFNet implements multi-branch convolution by grouped convolution as mentioned in
\cite{xie2017aggregated}, which has fewer parameters and computations than traditional
convolution, making it a lightweight option.
The fusion path, like the other two paths, is also vectorized using the SAV module.
Then, the feature vectors of the three paths are concatenated to produce the final feature vector $F_{final}$
and the identification loss ($L_F$) is computed using $F_{final}$.

In summary, the training loss function is denoted as the sum of three-path losses:
\begin{equation}
  \label{eq5}
  L = L_D + L_N + L_F,
\end{equation}

In the testing phase, we calculate the cosine similarity of samples in the probe and gallery,
using the final feature vector, $F_{final}$. More details of LDNFNet are given in Appendix A.

\section{Experiment}
In this section, we construct comparative analyses to verify the effectiveness of
our proposed DMDNet and LDNFNet. Specifically, we evaluate the performance of DMDNet
using classical denoising metrics such as Peak Signal-to-Noise Ratio (PSNR), Structural
Similarity Index Measure (SSIM), and Root Mean Square Error (RMSE), on the Bosphorus database.
We also measure the rank-one recognition accuracy of DMDNet on the Lock3DFace database.
Our method outperforms other denoising approaches, demonstrating superior results.
Furthermore, by combining DMDNet and LDNFNet, we achieve state-of-the-art recognition
performance on the Lock3DFace database.
Finally, extensive ablation experiments are implemented
to demonstrate the rationality of our proposed method.
\subsection{Datasets and Preprocessing}
Three high-quality datasets (FRGC v2, Bosphorus, BU3DFE)
and four low-quality datasets (Lock3DFace, USTC \cite{JiangZD20},
MultiSFace \cite{pini2021systematic}, IIIT-D \cite{goswami2013rgb}) are utilized for evaluation.

\textbf{FRGC v2} is collected by the laser 3D scanner and contains 4007 3D faces with
expression changes from 466 individuals.

\textbf{Bosphorus} is acquired by the high-precision structured-light 3D system and
consists of 4666 3D faces of 105 subjects
with expression, occlusion, and pose variations.

\textbf{BU3DFE} is captured with a 3D face imaging system and includes both
prototypical 3D facial expression shapes and 2D facial textures of 2500 faces from 100 subjects.

\textbf{Lock3DFace} is collected by the Kinect V2 camera,
composed of 5671 video sequences of 509 subjects and contains various variations
in expression, poses, time changes, and occlusions.

\textbf{USTC} is acquired using a PrimeSense camera within the same indoor environment,
comprising 24,839 RGB-D images depicting 873 distinct identities.

\textbf{MultiSFace} is collected by the Pmdtec Pico Flexx camera and
comprises upper-body recordings from 31 individuals, each with 16 unique sequences.

\textbf{IIIT-D} has 4603 depth maps of 106 subjects, which were captured by Kinect V1 with
moderate pose and expression variations.

We employ a uniform preprocessing pipeline proposed in \cite{mu2019led3d},
including nose tip based cropping, outlier removing, and hole filling.
In addition, each depth face is resized to $128 \times 128$ and normalized to a range of $[0,255]$. 
To overcome the risk of overfitting caused by the limited amount of 3D face data,
we implement the pose generation method proposed in
\cite{mu2019led3d,mu2021refining} for data augmentation.

\subsection{Settings}
\textbf{Depth face denoising.} 
Using the data quality degradation methods (downsampling and adding noise)
proposed in the 3D-FRM \cite{mu2021refining}, we generate the high- and low-quality
depth face pairs on the FRGC v2, BU3DFE, and Bosphorus datasets.
The first two are used for training of DMDNet,
and the last one is used for evaluating denoising metrics.
In total, the training dataset comprises 149,661 depth face pairs,
while the total testing dataset consists of 4,662 depth face pairs.
To compute the perceptual loss $L_{per}^{l_1}$,
we pre-train a face recognition model using all the high-quality data from
these three datasets. The architecture of the model is the same as Led3D \cite{mu2019led3d}.

We use the Adam optimizer with an initial learning rate of $10^{-4}$.
The model is trained for 100 epochs with a batch size of 64,
and the learning rate decays by a factor of 0.5 every 20 epochs.
The parameters ($\lambda_{1}$, $\lambda_{2}$, and $\lambda_{3}$) of the loss function
are set to 1, 0.5, and 0.001, respectively.

\textbf{Depth face recognition.} 
Lock3DFace is the largest publicly available low-quality 3D face database, and we apply
it in the experiment for face recognition in terms of rank-one accuracy.
We adopt the same protocol in \cite{mu2019led3d} in which the training and testing sets
are divided by subjects.
We use all data from 340 randomly selected subjects for training and the remaining data from 
169 subjects for testing.

To demonstrate that better face recognition can be achieved using our proposed denoising network DMDNet,
we employ the same settings as in 3D-FRM \cite{mu2021refining} for fair comparison.
Firstly, we use Led3D \cite{mu2019led3d} as the recognition model
and pre-train it with augmented low-quality depth face images from the training set of Lock3DFace.
Next, we use DMDNet to denoise all the low-quality data in Lock3DFace, and then fine-tune the
pre-trained model based on the denoised training data. Finally, we evaluate the recognition
accuracy of the denoised testing data.

When pre-training, we use the SGD optimizer with a learning rate of $10^{-2}$
for 100 epochs.
We fine-tune the pre-trained model using the SGD optimizer with a learning rate
of $5 \times 10^{-3}$ for 50 epochs.
The batch size is set to 384 for both pre-training and fine-tuning.

To compare further with the current state-of-the-art works,
we construct another experiment replacing the recognition model with our proposed
LDNFNet, with all other settings remaining unchanged.

\begin{table}[!t]
  \begin{center}
    \small
    \begin{tabular}{c c c c c c c c c c}
      \toprule
      \multirow{2}{*}{\textbf{Method}} & \multicolumn{3}{c}{\textbf{Bosphorus}} & \multicolumn{6}{c}{\textbf{Lock3DFace}} \\
      & PSNR$\uparrow$  & SSIM$\uparrow$ & RMSE$\downarrow$ & NU & FE & PS & OC & TM & Total \\
      \midrule
      Led3D & - & - & - & 99.62 & 97.62 & 64.81 & 68.93 & 64.97 & 81.02 \\[1.1ex]
      DDRNet & 31.44 & 95.96 & 0.0623 & 99.92 & 96.83 & 64.61 & 67.92 & 72.65 & 81.75 \\[1.1ex]
      3D-FRM & 32.04 & 96.35 & 0.0587 & 99.96 & 96.83 & 64.98 & 69.25 & \textbf{73.68} & 82.23 \\
      \midrule
      DMDNet & \textbf{32.60} & \textbf{97.31} & \textbf{0.0470} & \textbf{100} & \textbf{99.13} & \textbf{79.76} & \textbf{80.63} & 73.17 & \textbf{86.50} \\
      \bottomrule
      \multicolumn{10}{l}{NU: neutral. FE: expression. OC: occlusion. PS: pose. TM: time.}
    \end{tabular}
    \caption{
      Quantitative comparison results of different models for denoising metrics on
      Bosphorus database, and quantitative results in terms of rank-one recognition
      rate (\%) on the Lock3DFace database using the denoised depth faces generated
      by different denoising methods.
    }
    \label{tab1}
  \end{center}
\end{table}

\subsection{Results}\label{results}
\subsubsection{Results of DMDNet} 
We evaluate the performance of DMDNet from two perspectives: denoising metrics and rank-one recognition accuracy,
and compare it with other denoising works \cite{mu2021refining,yan2018ddrnet}.
The results are presented in Table \ref{tab1}. As observed,
our method outperforms other methods in terms of PSNR, SSIM, and RMSE, exhibiting an
improvement of 0.56db, 0.96\%, and
0.117, respectively, compared to 3D-FRM.
This demonstrates that our proposed DMDNet is more effective for depth face denoising.
Moreover, upon denoising the data in the Lock3DFace database using our method,
we can achieve higher face recognition accuracy on almost all the subsets,
indicating that DMDNet can maintain the original identity information and reconstruct
high-frequency facial details that are helpful for recognition.

\begin{table}[t!]
  \begin{center}
    \small
    \begin{tabular}{c c c c c c c c }
      \toprule
      \multirow{2}{*}{\textbf{Method}} & \multirow{2}{*}{\textbf{Input}} & \multicolumn{6}{c}{\textbf{Lock3DFace}} \\
      & & NU & FE & PS & OC & TM & Total \\
      \midrule
      VGG-16 \cite{simonyan2014very} & Depth & 99.57 & 94.76 & 49.21 & 44.68 & 34.50 & 70.58 \\[1.1ex]
      Inception-V2 \cite{ioffe2015batch} & Depth & 98.97 & 93.56 & 54.14 & 56.98 & 42.17 & 74.44 \\[1.1ex]
      ResNet-34 \cite{he2016deep} & Depth & 99.29 & 96.09 & 61.39 & 54.91 & 45.00 & 76.56 \\[1.1ex]
      MobileNet-V2 \cite{sandler2018mobilenetv2} & Depth & 98.91 & 95.74 & 69.92 & 61.44 & 43.00 & 79.49 \\[1.1ex]
      Cui et al. \cite{cui2018improving} & Depth & 99.55 & 98.03 & 65.26 & 81.62 & 55.79 & 79.85 \\[1.1ex]
      Led3D \cite{mu2019led3d} & Depth+Normal & 99.62 & 98.17 & 70.38 & 78.10 & 65.28 & 84.22 \\[1.1ex]
      MQFNet \cite{lin2021high} & Normal & \underline{99.95} & 97.31 & 73.61 & 80.97 & 61.67 & 86.55 \\[1.1ex]
      LMFNet \cite{zhao2022lmfnet} & Depth+Normal & \underline{99.95} & 99.21 & 79.05 & 84.40 & 76.31 & \underline{88.01} \\
      \midrule
      LDNFNet & Depth+Normal & \textbf{100} & \underline{99.37} & \underline{82.94} & \underline{85.39} & 71.95 & 87.71 \\
      LDNFNet* & Depth+Normal & \underline{99.95} & \textbf{99.49} & \textbf{83.98} & \textbf{85.49} & \textbf{76.32} & \textbf{88.94} \\
      \bottomrule
    \end{tabular}
    \caption{
      Comparison of the rank-one recognition rate (\%) with other works using the
      combination of DMDNet and LDNFNet. The Input column denotes the
      modality adopted by the network. In the last two lines,
      the result of LDNFNet* indicates training with denoised data,
      while the result of LDNFNet indicates training with original noisy data.
      The best result is shown in bold, and the second-best result is highlighted with an underline.
    }
    \label{tab2}
  \end{center}
\end{table}

\subsubsection{Results of LDNFNet} 
To expand upon comparisons with other 3D FR works, we replace Led3D with the proposed LDNFNet as the
recognition model.
The resulting average performance and accuracy on all subsets are presented in Table \ref{tab2}.
LDNFNet trained with original noisy data outperforms previous work on almost all subsets,
demonstrating its ability to fully utilize the information from the depth and normal maps,
which provides more discriminative features for face recognition when compared to Led3D.
Additionally, when using the denoised data from DMDNet for fine-tuning,
LDNFNet* achieves state-of-the-art results on all subsets,
and the average performance is also the highest.
This further proves the effectiveness of our denoising method.

\begin{figure}[!b]
 \centering 
 \includegraphics[width=5.0in]{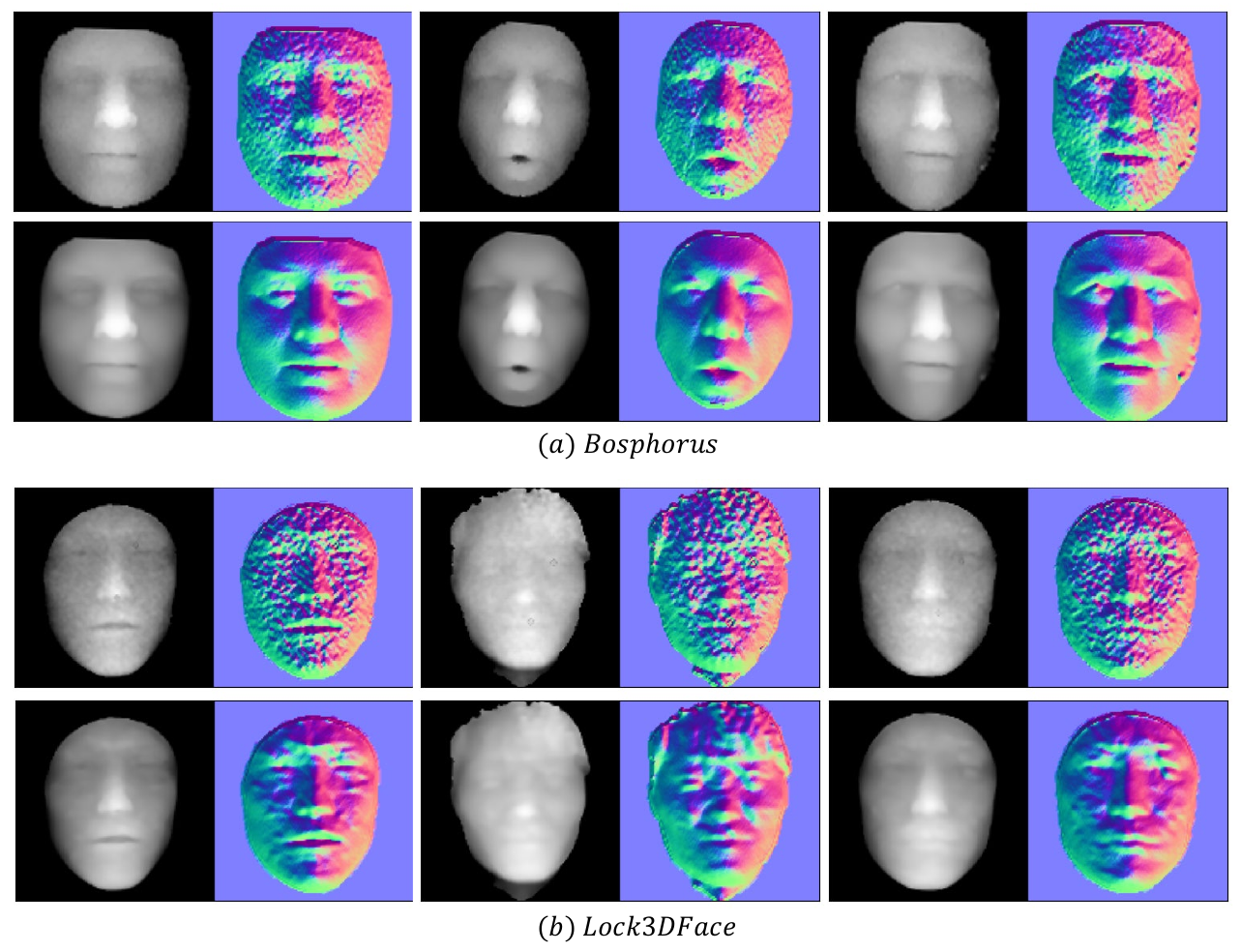}
 \caption{
  Demonstration of the denoised depth faces generated by DMDNet.
  The first and third rows are the original noisy depth faces,
  and the second and fourth rows are denoised depth faces.
  To show the effect of denoising more clearly, we also visualize the corresponding normal map.
  (a) The denoised results in the Bosphorus database.
  (b) The denoised results in the Lock3DFace database.
 }
 \label{fig5}
\end{figure}

\subsection{Model Analysis}
In this section, we showcase the denoising capabilities of DMDNet on the
Bosphorus database and Lock3DFace database.
Additionally, we demonstrate the generalization of DMDNet by evaluating
its performance on three databases acquired from different depth sensors.
To further validate the effectiveness of our proposed methods, a series of ablation experiments
are conducted.

\subsubsection{Visualization for Denoised Images}
We apply the trained DMDNet to denoise images from the Bosphorus and Lock3DFace databases.
The resulting denoised samples are displayed in Figure \ref{fig5}, showing that DMDNet effectively
reduces noise and generates smoother depth faces.
Due to the lack of real low- and high-quality data pairs, we use the synthetic data
pairs from the FRGC v2 and BU3DFE databases to train DMDNet.
Therefore, the trained DMDNet performs better on denoising
low-quality data synthesized from the Bosphorus database.
However, it also performs well on the real-world and noisy Lock3DFace database, indicating DMDNet's
robustness against real noise distributions and its potential for practical applications.

\begin{table}[!t]
    \centering
    \begin{subtable}[t]{0.45\textwidth}
        \centering
        \begin{tabular}{c@{\hspace{10pt}}c@{\hspace{10pt}}c}
            \toprule
            \textbf{Method} & \textbf{USTC} & \textbf{MultiSFace} \\
            \midrule
            Led3D & 73.69 & 59.34 \\
            Led3D* & \textbf{77.24} & \textbf{61.65} \\
            \bottomrule
        \end{tabular}
        \caption{
         Rank-one recognition rate (\%) on USTC and MultiSFace.        }
        \label{tab:USTC-MultiSFace}
    \end{subtable}
    \hfill
    \begin{subtable}[t]{0.45\textwidth}
        \centering
        \begin{tabular}{c@{\hspace{10pt}}c}
            \toprule
            \textbf{Method} & \textbf{Accuracy} \\
            \midrule
            Inception-V2 & 65.58 \\
            Led3D & 74.27 \\
            LMFNet & 81.13 \\
            \midrule
            LDNFNet & 86.54 \\
            LDNFNet* & \textbf{88.10} \\
            \bottomrule
        \end{tabular}
        \caption{
          Compared with the rank-one recognition rate (\%) on the IIIT-D database.
        }
        \label{tab:IIITD}
    \end{subtable}
    \caption{
      Generalization evaluation results on different low-quality 3D face datasets.
      Methods with * indicate training with denoised data, while those without * 
      indicate training with original noisy data.
    }
    \label{tab:all_datasets}
\end{table}

\subsubsection{Generalization Evaluation}
In section \ref{results}, we demonstrate the effectiveness of DMDNet and LDNFNet on the
Lock3DFace database. To further validate the generalization of our methods to
different types of noise and quality variations,
we conduct denoising and recognition experiments on three additional databases: USTC, MultiSFace, and IIIT-D.
These databases were collected using different consumer-grade depth sensors, namely PrimeSense, Pico Flexx, and Kinect V1.

Specifically, we adopt the evaluation method mentioned in Section \ref{results}
to assess the face recognition rate before and after denoising using DMDNet on these databases.
For the USTC and MultiSFace databases, we employ Led3D as our recognition
model due to the lack of suitable comparisons.
The training set consists of randomly selected 70 percent of the subjects,
while the remaining 30 percent form the test set.
Regarding the IIIT-D database, we follow the same training and testing settings as \cite{zhao2022lmfnet}
and employ LDNFNet to obtain the best results for comparison with other works.

The results are shown in Table \ref{tab:all_datasets}.
Across all databases, the recognition results after denoising demonstrate improvement,
confirming the good generalization of our DMDNet.
Additionally, on the IIIT-D database, our proposed LDNFNet exhibits significant advantages over
other methods, irrespective of denoising, thus demonstrating the superiority of LDNFNet.

\begin{table}[!t]
  \setlength{\tabcolsep}{2.1pt}
  \begin{center}
    \small
    \begin{tabular}{c c c c c c c c c c c c}
      \toprule
      \multicolumn{3}{c}{\textbf{Loss}} & \multicolumn{3}{c}{\textbf{Bosphorus}} & \multicolumn{6}{c}{\textbf{Lock3DFace}} \\
      $L_{rec}^{l_1}$ & $L_{rec}^{SSIM}$ & $L_{per}^{l_1}$ & PSNR$\uparrow$ & SSIM$\uparrow$ & RMSE$\downarrow$ & NU & FE & PS & OC & TM & Total \\
      \midrule
      \checkmark & - & - & 32.39 & 97.17 & 0.0481 & \textbf{100} & 98.97 & 79.17 & 80.58 & 72.42 & 86.18 \\[1.1ex]
      \checkmark & \checkmark & - & \textbf{32.63} & \textbf{97.32} & \textbf{0.0468} & \textbf{100} & \textbf{99.33} & \textbf{79.81} & 80.63 & 71.20 & 86.20 \\[1.1ex]
      \checkmark & \checkmark & \checkmark & 32.60 & 97.31 & 0.0470 & \textbf{100} & 99.13 & 79.76 & \textbf{81.13} & \textbf{73.17} & \textbf{86.50} \\
      \bottomrule
    \end{tabular}
    \caption{
      Ablation study on the composition of the loss functions of DMDNet.
    }
    \label{tab3}
  \end{center}
\end{table}

\subsubsection{Ablation Study for DMDNet}
We first analyze the composition of the loss function of DMDNet.
We individually use the losses as $L_{rec}^{l_1}$, $L_{rec}^{l_1}+L_{rec}^{SSIM}$,
and $L_{rec}^{l_1}+L_{rec}^{SSIM}+L_{per}^{l_1}$ to guide the training of DMDNet.
We then follow the same settings to test the denoising metrics on the Bosphorus database and
the face recognition accuracy on the Lock3DFace database.
The testing results for the three losses are shown in Table \ref{tab3}.
As we can see, compared to using only L1 loss $L_{rec}^{l_1}$ as the reconstruction loss,
the joint use of L1 loss $L_{rec}^{l_1}$ and SSIM loss $L_{rec}^{SSIM}$ can effectively improve
the denoising metrics and slightly improve face recognition performance.
In addition, although the use of perceptual loss $L_{per}^{l_1}$ will cause a
certain decrease in denoising metrics, it helps preserve facial identity information
and face recognition.
We argue that under the tradeoff between denoising performance and recognition accuracy,
introducing perceptual loss is worthwhile.

\begin{table}[!t]
  \setlength{\tabcolsep}{2.4pt}
  \begin{center}
    \small
    \begin{tabular}{c c c c c c c c c c c c}
      \toprule
      \multicolumn{3}{c}{\textbf{Method}} & \multicolumn{3}{c}{\textbf{Bosphorus}} & \multicolumn{6}{c}{\textbf{Lock3DFace}} \\
      $e_{ff}$ & $e_{ce}$ & MSDF & PSNR$\uparrow$ & SSIM$\uparrow$ & RMSE$\downarrow$ & NU & FE & PS & OC & TM & Total \\
      \midrule
      - & - & - & 32.47 & 97.16 & 0.0477 & \textbf{100} & \textbf{99.25} & 78.97 & 81.08 & 72.54 & 86.32 \\[1.1ex]
      \checkmark & - & - & 32.49 & 97.19 & 0.0476 & 99.95 & 99.17 & \textbf{80.01} & 80.48 & 71.87 & 86.23 \\[1.1ex]
      - & \checkmark & - & 32.48 & 97.18 & 0.0477 & \textbf{100} & 99.21 & 79.32 & 80.93 & 72.22 & 86.28 \\[1.1ex]
      - & - & \checkmark & 32.54 & 97.29 & 0.0473 & \textbf{100} & 99.05 & \textbf{80.01} & 80.88 & 72.50 & 86.42 \\[1.1ex]
      \checkmark & \checkmark & - & 32.50 & 97.20 & 0.0476 & \textbf{100} & 99.17 & 79.91 & \textbf{81.18} & 72.46 & 86.48 \\[1.1ex]
      \checkmark & \checkmark & \checkmark & \textbf{32.60} & \textbf{97.31} & \textbf{0.0470} & \textbf{100} & 99.13 & 79.76 & 80.63 & \textbf{73.17} & \textbf{86.50} \\
      \bottomrule
    \end{tabular}
    \caption{
      Ablation study for positional encoding and MSDF.
      $e_{ff}$ denotes the Fourier Features and $e_{ce}$ denotes the Coordinate Embeddings.
    }
    \label{tab4}
  \end{center}
\end{table}

To demonstrate the effectiveness of the two strategies (Positional Encoding and MSDF)
proposed in the decoding function, we evaluate the impact of each component on denoising performance
and face recognition accuracy.
The results are shown in Table \ref{tab4}. Each component used alone improves the denoising performance
of DMDNet, and their combined use further enhances the performance of face recognition.
In particular, the inclusion of MSDF significantly improves the denoising metrics,
with PSNR and SSIM increasing by 0.1db and 0.11\%, respectively,
and RMSE decreasing by 0.0006.
Furthermore, the use of MSDF yields the highest face recognition accuracy, which highlights its crucial role.

\begin{figure}[!h]
 \centering 
 \includegraphics[width=5.0in]{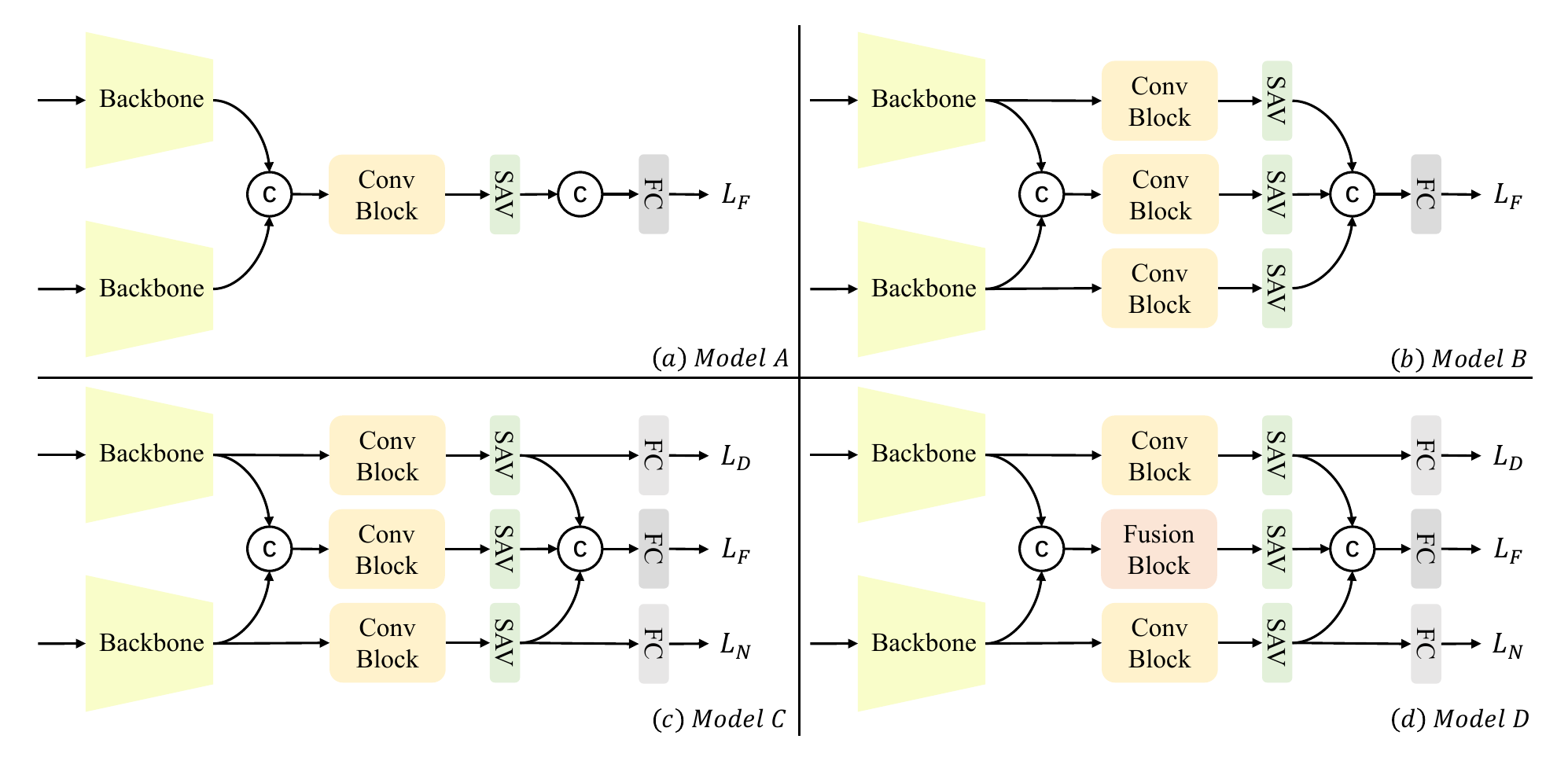}
 \caption{
  Four model structure diagrams are constructed for the LDNFNet ablation study.
 }
 \label{fig6}
\end{figure}

\begin{table*}[!h]
  \begin{center}
    \begin{tabular}{c@{\hspace{20pt}}c@{\hspace{20pt}}c@{\hspace{20pt}}c@{\hspace{20pt}}c@{\hspace{20pt}}c@{\hspace{20pt}}c }
      \toprule
      \multirow{2}{*}{\textbf{Method}} & \multicolumn{6}{c}{\textbf{Lock3DFace}} \\
      & NU & FE & PS & OC & TM & Total \\
      \midrule
      Model A & 99.85 & 98.54 & 79.32 & 83.68 & 66.67 & 85.32 \\[1.1ex]
      Model B & \textbf{100} & 98.82 & 79.32 & 82.18 & 70.02 & 85.91 \\[1.1ex]
      Model C & 99.95 & 99.05 & 81.35 & \textbf{85.39} & \textbf{72.77} & 87.32 \\[1.1ex]
      Model D & \textbf{100} & \textbf{99.37} & \textbf{82.94} & \textbf{85.39} & 71.95 & \textbf{87.71} \\
      \bottomrule
    \end{tabular}
    \caption{
      Comparison of the rank-one recognition rates of four models on
      Lock3DFace database. Model D is the ultimate scheme of our LDNFNet.
    }
    \label{tab5}
  \end{center}
\end{table*}

\subsubsection{Ablation Study for LDNFNet}
In this section, we conduct an ablation study for LDNFNet by constructing
three contrasting models (Models A, B, and C) based on the structure of LDNFNet (Model D),
as illustrated in Figure \ref{fig6}.
The aim of this study is to validate the effectiveness of our LDNFNet design.
Each model removes or replaces specific components from LDNFNet, allowing
us to analyze their impact on performance.

Specifically, Model A is the simplest variant, comprising only one fusion
path, and it employs a ConvBlock as the fusion module for depth and normal
features. Model B is an extension of Model A, incorporating three paths but
with only one fusion loss, denoted as $L_{F}$. Model C further includes
two additional auxiliary losses, $L_{D}$ and $L_{N}$, in addition to the
fusion loss $L_{F}$. Finally, Model D represents the ultimate scheme of our
LDNFNet, where the fusion module in Model C is replaced with a multi-branch
fusion block.
To evaluate the performance of these four models, we train them using the
training set of the Lock3DFace database and assess their rank-one accuracy
on the testing set.

\begin{table}[!t]
  \begin{center}
    \begin{tabular}{c@{\hspace{20pt}}c@{\hspace{20pt}}c}
      \toprule
      \textbf{Block} & \textbf{Params(M)} & \textbf{MAdds(M)} \\
      \midrule
      ConvBlock & 8.29 & 539.14 \\
      Multi-branch Fusion Block & \textbf{0.99} & \textbf{65.11} \\
      \bottomrule
    \end{tabular}
    \caption{
      Comparison of the number of parameters and
      multiply-adds (MAdds) for ConvBlock and Fusion Block.
    }
    \label{tab6}
  \end{center}
\end{table}

Table \ref{tab5} presents the results of our evaluation.
It is evident from the table that the three-path structure (Model B)
outperforms the single-path structure (Model A), indicating the advantage
of incorporating multiple paths in LDNFNet.
Furthermore, the inclusion of auxiliary losses (Model C) for the depth and
normal paths is crucial in achieving improved results.
Additionally, replacing the ConvBlock with the more sophisticated multi-branch
fusion block (Model D) leads to a performance improvement, emphasizing the
efficacy of our design choices in LDNFNet.

Moreover, we provide a comprehensive comparison of the parameters and operations
between the ConvBlock and the multi-branch fusion block.
The number of operations is measured in multiply-adds (MAdds), and the results are presented in Table \ref{tab6}.
Significantly, the multi-branch fusion block requires fewer parameters and computations
compared to the ConvBlock, which aligned with our lightweight design goal
and supports the feasibility of deploying LDNFNet in resource-constrained environments.

\section{Conclution}
In this paper, we propose a novel Depth Map Denoising Network (DMDNet)
based on the Denoising Implicit Image Function (DIIF) to enhance the quality of facial depth
images for low-quality 3D face recognition by eliminating noise.
Moreover, we introduce two effective strategies in the decoding stage: positional encoding and multi-scale decoding fusion.
These approaches significantly improve the denoising performance of DMDNet.
We evaluate denoising metrics on the Bosphorus database and assess face recognition accuracy
on the Lock3DFace database. The results indicate that our proposed DMDNet outperforms other methods.
Additionally, to achieve better 3D FR performance using the denoised images,
we design a Lightweight Depth and Normal
Fusion Network (LDNFNet), which leverages multiple modalities (i.e., depth and normal images)
through a multi-branch fusion block.
By combining DMDNet and LDNFNet, we achieve state-of-the-art results on the Lock3DFace database.
Furthermore, experimental results from databases obtained from three different depth sensors
demonstrate the robustness of our proposed methods,
and a series of ablation experiments also confirm the rationality of our methods.

Our work is the first to introduce implicit neural representations to the 3D face denoising task.
Despite the promising performance, our model is relatively simple in its design of the decoding function,
which limits the flexibility of the model.
In future work, we will explore designing more sophisticated decoding functions
to further improve the denoising performance.

\section*{Acknowledgment}
This work was funded by the Beijing University of Posts and
Telecommun-ications-China Mobile Research Institute Joint Innovation Center,
National Natural Science Foundation of China under Grant No. 62236003,
and also by China Postdoctoral Science Foundation under Grant 2022M720517. 

\bibliographystyle{elsarticle-num} 
\bibliography{ref}

\newpage

\section*{Appendix A: Detailed Structure of LDNFNet}
\label{Appendix:A}
\begin{figure}[htbp]
 \centering 
 \includegraphics[width=5.0in]{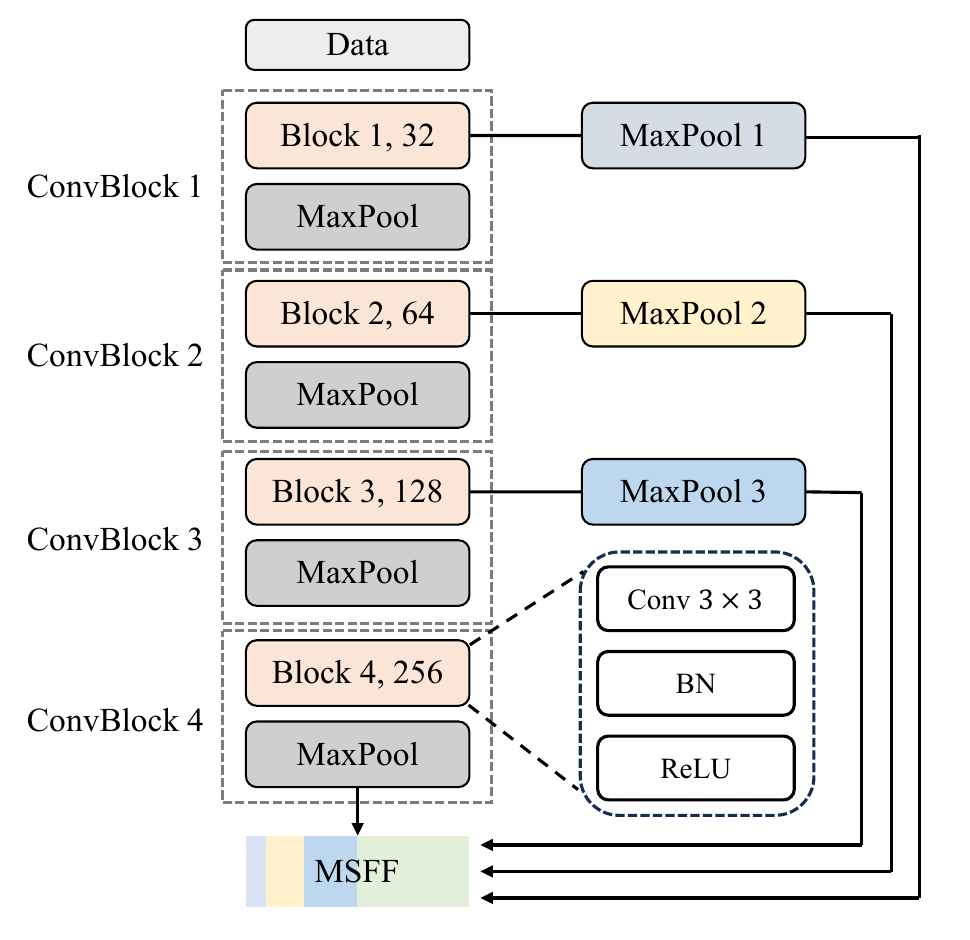}
 \caption{
  The specific architecture of the LDNFNet backbone network.
 }
 \label{fig:LDNFNet_backbone}
\end{figure}

This section presents the network details of LDNFNet to facilitate replication by readers.
As illustrated in Figure \ref{fig:LDNFNet}, our proposed LDNFNet comprises four main modules:
the Backbone network, ConvBlock, FusionBlock, and SAV.
The Backbone network replicates the structure of Led3D \cite{mu2019led3d},
which is depicted in Figure \ref{fig:LDNFNet_backbone}.
For specific parameter configurations and output dimensions of each module, please refer to Table \ref{tab:ldnfnet}.

\begin{table}[htbp]
\renewcommand{\arraystretch}{0.6}
\centering
\begin{tabular}{c|c|c|c|c}
\hline
\textbf{Module} & \textbf{Block} & \textbf{Layer} & \textbf{Parameters} & \textbf{Output Size} \\
\hline
\multirow{17}{*}{Backbone}
  & \multirow{4}{*}{ConvBlock 1} & Conv & $3,1,1,1$ & \multirow{3}{*}{$32 \times 128 \times 128$} \\
  & & BN & - & \\
  & & ReLU & - & \\
  \cline{3-5}
  & & MaxPool & $3,2,1,-$ & $32 \times 64 \times 64$ \\
  \cline{2-5}
  & \multirow{4}{*}{ConvBlock 2} & Conv & $3,1,1,1$ & \multirow{3}{*}{$64 \times 64 \times 64$} \\
  & & BN & - & \\
  & & ReLU & - & \\
  \cline{3-5}
  & & MaxPool & $3,2,1,-$ & $64 \times 32 \times 32$ \\
  \cline{2-5}
  & \multirow{4}{*}{ConvBlock 3} & Conv & $3,1,1,1$ & \multirow{3}{*}{$128 \times 32 \times 32$} \\
  & & BN & - & \\
  & & ReLU & - & \\
  \cline{3-5}
  & & MaxPool & $3,2,1,-$ & $128 \times 16 \times 16$ \\
  \cline{2-5}
  & \multirow{4}{*}{ConvBlock 4} & Conv & $3,1,1,1$ & \multirow{3}{*}{$256 \times 16 \times 16$} \\
  & & BN & - & \\
  & & ReLU & - & \\
  \cline{3-5}
  & & MaxPool & $3,2,1,-$ & $256 \times 8 \times 8$ \\
  \cline{2-5}
  & \multirow{4}{*}{MSFF} & MaxPool 1 & $33,16,16,-$ & $32 \times 8 \times 8$ \\
  & & MaxPool 2 & $17,8,8,-$ & $64 \times 8 \times 8$ \\
  & & MaxPool 3 & $9,4,4,-$ & $128 \times 8 \times 8$ \\
  & & Concate & - & $480 \times 8 \times 8$ \\
  \hline
\multirow{3}{*}{ConvBlock}
  & \multirow{3}{*}{ConvBlock 6} & Conv & $3,1,1,1$ & \multirow{3}{*}{$480 \times 8 \times 8$} \\
  & & BN & - & \\
  & & ReLU & - & \\
  \hline
\multirow{9}{*}{FusionBlock}
  & \multirow{3}{*}{ConvBlock 7} & Conv & $1,1,0,1$ & \multirow{3}{*}{$480 \times 8 \times 8$} \\
  & & BN & - & \\
  & & ReLU & - & \\
  \cline{2-5}
  & \multirow{3}{*}{ConvBlock 8} & Conv & $3,1,1,\textbf{32}$ & \multirow{3}{*}{$480 \times 8 \times 8$} \\
  & & BN & - & \\
  & & ReLU & - & \\
  \cline{2-5}
  & \multirow{3}{*}{ConvBlock 9} & Conv & $1,1,0,1$ & \multirow{3}{*}{$960 \times 8 \times 8$} \\
  & & BN & - & \\
  & & ReLU & - & \\
  \hline
\multirow{2}{*}{SAV}
  & \multirow{2}{*}{SAV} & Conv & $8,8,0,N$ & $N \times 1 \times 1$ \\
  & & Flatten & - & $N$ \\
\hline
\end{tabular}
\caption{
  Detailed Structure of LDNFNet.
  The \textbf{Parameters} column goes from left to right: kernel size, stride, padding, and groups.
  The parameter N in the SAV module is equal to the number of input channels.
  The \textbf{Output Size} column indicates $Channels \times Height \times Width$.
  }
\label{tab:ldnfnet}
\end{table}

\section*{Appendix B: Discussion on the Encoder Design of DMDNet}
In this section, we present a detailed explanation of the encoder design for our DMDNet.
As discussed in Section \ref{Encoder}, our encoder is primarily constructed by stacking a series of
ResBlocks and DS-ResBlocks. The ResBlock is a fundamental module originally proposed in EDSR \cite{lim2017enhanced}
for low-level image tasks, such as super-resolution and denoising.
Notably, the ResBlcok in EDSR differs from the residual block in the original ResNet \cite{he2016deep}
as it eliminates the batch normalization layer after each convolutional layer.
This modification reduces the model's memory consumption and enhances its flexibility.
For a more comprehensive understanding of the ResBlock design, please refer to \cite{lim2017enhanced}.
Due to the excellent performance of ResBlock on low-level image tasks,
we adopt it as the base module of our encoder.
To further optimize the encoder, we introduce a small change to the ResBlock by adjusting
the stride of the first convolutional layer to 2.
This adjustment downsamples the feature map, thereby reducing both memory consumption and
computation overhead. We refer to this modified module as DS-ResBlock.

Another crucial design aspect of our encoder involves the utilization of both the
depth map and its corresponding normal map. This approach is supported by existing
works \cite{mu2019led3d,mu2021refining}, which demonstrate that normal maps provide
complementary information to enhance task performance.
Referring to 3D-FRM \cite{mu2021refining}, we perform an early fusion of features
extracted from the depth map and normal map.
Our experiments substantiate that employing the normal map modality for assistance
yields superior results compared to using only the depth map modality.
We present the comparison results in Table \ref{tab:appendixB}.

\begin{table}[htbp]
  \begin{center}
    \begin{tabular}{c c c c}
      \toprule
      \multirow{2}{*}{\textbf{Input}} & \multicolumn{3}{c}{\textbf{Bosphorus}} \\
      & PSNR$\uparrow$ & SSIM$\uparrow$ & RMSE$\downarrow$ \\
      \midrule
      Depth & 32.52 & 97.28 & 0.0474 \\
      Depth+Normal & \textbf{32.60} & \textbf{97.31} & \textbf{0.0470} \\
      \bottomrule
    \end{tabular}
    \caption{
    Comparison of denoising performance for different inputs.
    }
    \label{tab:appendixB}
  \end{center}
\end{table}

Despite using a relatively simple fusion approach,
the denoising performance of DMDNet has been significantly enhanced.
We acknowledge that employing more sophisticated fusion techniques could
lead to even greater performance improvements.
However, such investigations are beyond the primary focus of our work,
and we refrain from delving extensively into this aspect.

\section*{Declaration of interests}
$\surd $ The authors declare that they have no known competing financial interests or personal relationships that could have appeared to influence the work reported in this paper.

$\Box$ The authors declare the following financial interests/personal relationships which may be considered as potential competing interests:

\newpage   
\section*{Author Biographies}

\textbf{Ruizhuo Xu} is a graduate student studying for a master's degree in
School of Artificial Intelligence, Beijing University of Posts and
Telecommunication, China.

\textbf{Ke Wang} received the M.S. degree from the Beijing Institute of
Technology in 2018. He is currently an algorithm engineer at China Mobile
Research Institute. His research interests include 3D reconstruction,
3D face recognition and multi-modal fusion.

\textbf{Chao Deng} received the M.S. degree and the Ph.D. degree from
Harbin Institute of Technology, Harbin, China, in 2003 and 2009 respectively.
He is currently a deputy general manager with AI center of China Mobile Research Institute.
His research interests include machine learning and artificial intelligence for ICT operations.

\textbf{Mei Wang} received the M.S. degree and Ph.D. degree in
information and communication engineering from Beijing University of Posts
and Telecommunications (BUPT), Beijing, China, in 2013 and 2022, respectively.
She is currently a Postdoc in the School of Artificial Intelligence,
Beijing University of Posts and Telecommunications, China.
Her research interests include computer vision, with a particular emphasis in face recognition, domain adaptation and AI fairness.

\textbf{Xi Chen} received her M.S. degree from Communication University of China.
She works as an algorithm engineer in image and video processing
and computer vision at China Mobile Research Institute since 2018.

\textbf{Wenhui Huang} received her M.S. degree from Beijing University of
Post and Telecommunication in 2005. She worked as an algorithm engineer in
image and video processing, and computer vision, for more than 16 years.
She is currently an algorithm engineer at China Mobile Research Institute,
working on the industrial application of computer vision.

\textbf{Junlan Feng}, IEEE Fellow.
Dr.Feng received her Ph.D. on Speech Recognition from Chinese Academy of Sciences in 2001.
She had been a principal researcher at AT\&T Labs Research and has been the chief scientist of China Mobile Research since 2013.

\textbf{Weihong Deng} received the B.E. degree in information engineering
and the Ph.D. degree in signal and information processing from the
Beijing University of Posts and Telecommunications (BUPT), Beijing, China,
in 2004 and 2009, respectively.
He is currently a professor in School of Artificial Intelligence, BUPT.
His research interests include trustworthy biometrics and affective computing,
with a particular emphasis in face recognition and expression analysis.

\end{document}